\documentclass[11pt]{article}
\usepackage{colacl} 
\usepackage{epsfig}

\title{Explaining away ambiguity: Learning verb selectional preference
with Bayesian networks\thanks{We would like to thank the Brown
Laboratory for Linguistic Information Processing; Thomas Hofmann; Elie 
Bienenstock; Philip Resnik, who
provided us with training and test data; and Daniel Garcia for his
help with the SMILE library of
classes for Bayesian networks that we used for our experiments. This
research was supported by NSF awards 9720368, 9870676 
and 9812169.}} 

\author{Massimiliano Ciaramita \and Mark Johnson \\ Cognitive and
Linguistic Sciences \\ Box 1978, Brown University \\ Providence, RI
02912, USA \\ {\tt massimiliano\_ciaramita@brown.edu}\hspace{.25in}{\tt
mj@cs.brown.edu}}  

\begin{document}
\maketitle
\bibliographystyle{acl}
\begin{abstract}
This paper presents a Bayesian model for unsupervised learning of verb
selectional preferences. For each verb the model creates a Bayesian
network whose architecture is determined by the lexical hierarchy of
Wordnet and whose parameters are estimated from a list of verb-object
pairs found from a corpus. ``Explaining away'', a well-known property of
Bayesian networks, helps the model deal in a natural fashion with word
sense ambiguity in the 
training data. On a word sense disambiguation test
our model performed better than other state of the art systems for
unsupervised learning of selectional preferences.
Computational complexity problems, ways of improving this
approach and methods for implementing ``explaining away'' in other
graphical frameworks are discussed.
\end{abstract}

\section{Selectional preference and sense ambiguity}
Regularities of a verb with respect to the semantic class of its
arguments (subject, object and indirect object) are called
\textbf{selectional preferences} (SP)~\cite{KF64,Chomsky65,JL83}.
The verb \textit{pilot} carries the information that
its object will likely be some kind of \textit{vehicle}; subjects of the verb
\textit{think} tend to be \textit{human}; and subjects of the verb 
\textit{bark} tend to be \textit{dogs}. For the sake of simplicity we
will focus on the verb-object relation although the 
techniques we will describe can be applied to other verb-argument
pairs.

Models of the acquisition of SP
are important in their own right and have applications in Natural
Language Processing (NLP). The selectional preferences of a verb can
be used to infer the possible meanings of an unknown argument of a
known verb; e.g., it might be possible to infer that 
\textit{xxxx} is a kind of dog from the following sentence: 
``The \emph{xxxx} barked all night''.
In parsing a sentence selectional preferences can be used to rank competing
parses, providing a partial measure of semantic
well-formedness. Investigating SP might help
us to understand the structure of the mental lexicon.
\begin{figure}
\begin{center}
\epsfig{file = 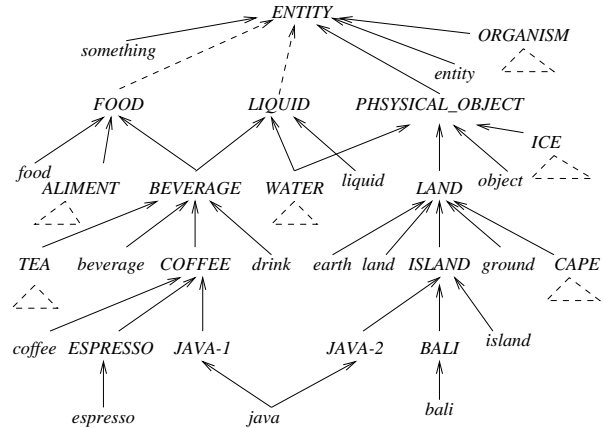, width = 3.1in}
\caption{A portion of Wordnet.}
\end{center}
\end{figure}

Systems for unsupervised learning of SP usually
combine statistical and knowledge-based approaches. The knowledge-base
component is typically a database that groups words into classes. In
the models we will see, the knowledge base is
Wordnet~\cite{Miller90}. Wordnet groups nouns into classes of synonyms 
representing concepts,
called \textbf{synsets}, e.g., \(\{car, auto,
automobile,\ldots\}\). 
A noun that belongs to several synsets is \textbf{ambiguous}.
A transitive and asymmetrical relation,
\textbf{hyponymy}, is defined between synsets. A synset 
is a hyponym of another synset if the former has the latter as a broader 
concept; for example, \textit{BEVERAGE} is a hyponym of
\textit{LIQUID}. Figure~1 depicts a portion of the hierarchy.

The statistical component consists of
predicate-argument pairs extracted from a corpus in which the
semantic class of the words is not indicated.
A trivial algorithm might get a list of words that occurred 
as objects of the verb and output the semantic classes the words
belong to according to Wordnet. For example,
if the verb \textit{drink} occurred with \textit{water} and $water \in
LIQUID$, the  
model would learn that \textit{drink} selects for \textit{LIQUID}. 
As~\newcite{Resnik97} and~\newcite{Abney99} have found, the main
problem these systems face is the presence of ambiguous words in the training
data. If the word \textit{java} also occurred as an object of
\textit{drink}, since
$java \in BEVERAGE$ and $java \in ISLAND$, this model would learn that
\textit{drink} selects for both $BEVERAGE$ and $ISLAND$. 

More complex models have been proposed. These models, though,
deal with word sense ambiguity by applying an \textit{unselective}
strategy similar to the one above; i.e., they assume that
ambiguous words provide equal evidence for all their senses.
These models choose as the concepts the verb selects for
those that are in common among several words (e.g., \textit{BEVERAGE} 
above).
This strategy works to the extent that these overlapping senses are
also the concepts the verb selects for.

\section{Previous approaches to learning selectional preference}

\subsection{Resnik's model}
Ours system is closely related to those proposed in~\cite{Resnik97}
and~\cite{Abney99}. The fact that a predicate \textit{p} selects for a
class \textit{c}, given a syntactic relation \textit{r}, can be
represented as a relation, \(selects(p,r,c)\);  
e.g., that \textit{eat} selects for \textit{FOOD} in object position
can be represented as \(selects(eat,object,FOOD)\). In~\cite{Resnik97}
selectional 
preference is quantified by comparing the prior distribution of a
given class \textit{c} appearing as an argument, \(P(c)\), and the
conditional probability of the same class given a predicate and a
syntactic relation \(P(c|p,r)\), e.g., \(P(FOOD)\) and
\(P(FOOD|eat,object)\). 
The relative entropy between \(P(c)\) and
\(P(c|p,r)\) measures how much the predicate constrains its arguments:
\begin{equation}
%\begin{eqnarray} \label{resnik1}
S(p,r)=D(P(c|p,r)~||~P(c))
%        &=& \sum_{c}~P(c|p,r)~\log {P(c|p,r) \over P(c)}\nonumber
%\end{eqnarray}
\end{equation}
Resnik defines the \textbf{selectional association} 
of a predicate for a particular class
\textit{c} to be the
portion of the 
selectional preference strength due to that class:
\begin{equation} \label{resnik2}
 A(p,r,c) = 
 {1 \over S(p,r)}~P(c|p,r)~\log{P(c|p,r) \over P(c)}
\end{equation}

Here the main problem is the estimation of \(P(c|p,r)\).
Resnik suggests as a plausible estimator
$\hat{P}(c|p,r) \stackrel{\mathrm{def}}{=}freq(p,r,c)/freq(p,r).$
But since the model is trained on data that are not
sense-tagged, there is no obvious way to estimate \(freq(p,r,c)\).
Resnik suggests considering each observation of a word as evidence for
each of the classes the word belongs to,
\begin{equation} \label{resnik3}
 freq(p,r,c) \approx \sum_{w \in c}~{count(p,r,w) \over classes(w)}
\end{equation}
where \(count(p,r,w)\) is the number of times the word \textit{w} occurred
as an argument of \textit{p} in relation \textit{r}, and
\(classes(w)\) is the number of classes \textit{w} belongs to.
\begin{figure}
\begin{center}
\epsfig{file=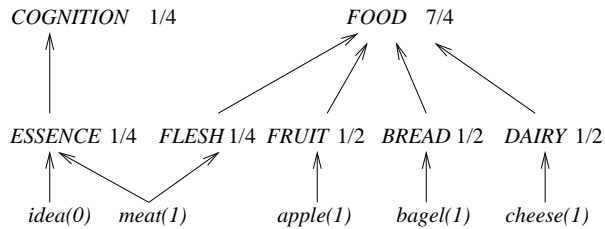, width = 3.1in}
\caption{Simplified Wordnet. The numbers next to the synsets
represent the values of $freq(p,r,c)$ estimated using (3),
the numbers in parentheses represent the values of $freq(p,r,w)$.}
\end{center}
\end{figure}
For example, suppose the system is trained on (\textit{eat,object}) 
pairs and the verb occurred once each with \textit{meat, apple,
bagel}, and  \textit{cheese}, and Wordnet is simplified as in
Figure~2. An ambiguous word like \textit{meat} provides
evidence also for classes that appear unrelated to those selected by
the verb. 
Resnik's assumption is that only the classes selected by the verb will 
be associated with each of the observed words, and hence will receive
the highest values for $P(c|p,r)$.
Using (3) we find that the highest frequency is in fact associated
with \textit{FOOD}: \(freq(eat,object,food)\approx {1\over 4}+{1\over
2}+{1\over 2}+{1\over 2} = {7\over 4}\) and \(P(FOOD|eat)=0.44\).
However, some evidence is found also for \textit{COGNITION}:
\(freq(eat,object,cognition) \approx {1\over 4}\) and
\(P(COGNITION|eat)=0.06\).

\subsection{Abney and Light's approach}
\newcite{Abney99} pointed out that the distribution of senses of an
ambiguous word is not uniform.
They noticed also that it is not clear how the probability 
\(P(c|p,r)\) is to be interpreted since there is no explicit
stochastic generation model involved.

They proposed a system that associates a 
Hidden Markov Model (HMM) with each predicate-relation pair
(\textit{p},\textit{r}).
Transitions between synset states represent the hyponymy relation, and
$\varepsilon$, the empty word, is emitted with probability 1;
transitions to a final state emit a word \textit{w} with probability
\(0\leq P(w) \leq 1\).
Transition and emission probabilities are estimated using the EM
algorithm on training data that consist of the nouns that occurred
with the verb. Abney and Light's model estimates $P(c|p,r)$ from the
model trained for $(p,r)$; the distribution $P(c)$ can be calculated
from a model trained for all nouns in the corpus.

\begin{figure}
\begin{center}
\epsfig{file=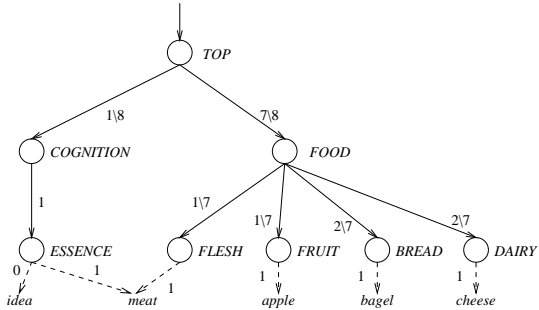, width = 2.8in}
\caption{The HMM version of the simple example.}
\end{center}
\end{figure}

This model did not perform as well as
expected. An ambiguous word in the model can be generated
by more than one state sequence. Abney and Light discovered that
the EM algorithm finds parameter values that associate some probability
mass with all the transitions in the multiple paths that lead
to an ambiguous word.
In other words, when there are several state sequences for the same word,
EM does not select one of them over the others.\footnote{As a matter of
fact, for this HMM there are (infinitely) many parameter
values that maximize the likelihood of the training data; i.e., the
parameters are not identifiable. The intuitively correct solution is
one of them, but so are infinitely many other, intuitively incorrect
ones. Thus it is no surprise that the EM algorithm cannot find the
intuitively correct solution.} Figure~3 shows the parameters estimated 
by EM for the same example as above. The transition 
to the \textit{COGNITION} state has been assigned a probability of
$1/8$ because it is part of a possible path to \textit{meat}. The HMM
model does not solve the problem of the unselective distribution of the
frequency of occurrence of an ambiguous word to all its senses. Abney and
Light claimed that this is a serious problem, particularly when the
ambiguous word is a frequent one, and caused the model to learn the
wrong selectional preferences.
To correct this undesirable outcome they introduced some smoothing and
balancing techniques. However, even with these modifications their
system's performance was below that achieved by Resnik's. 

\section{Bayesian networks}

A \textbf{Bayesian network}~\cite{Pearl88}, or Bayesian belief network
(BBN), consists of a set of \textbf{variables} and a set of
\textbf{directed edges} connecting the variables. 
The variables and the edges define a directed acyclic graph
(DAG) where each variable is represented by a node. Each variable is
associated with a finite number of (mutually exclusive) states. To
each variable \(A\) with parents \(B_1,...,B_n\) is attached a
\textit{conditional probability table} (CPT) \(P(A|B_1,...,B_n)\).
Given a BBN, Bayesian inference can be used to estimate
\textbf{marginal} and \textbf{posterior probabilities} given the
evidence at hand and the information stored in the CPTs, the
\textbf{prior probabilities}, by means of Bayes' rule,
\(P(H|E) = {P(H)P(E|H)\over P(E)}\), 
where H stands for hypothesis and E for evidence.
\begin{figure}
\begin{center}
\epsfig{file=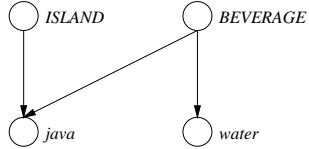, width = 1.6in}
\caption{A Bayesian network for word ambiguity.}
\end{center}
\end{figure}

Bayesian networks display an extremely interesting property called
\textbf{explaining away}. Word sense ambiguity in the process of
learning SP defines a problem that might be solved
by a model that implements an explaining away strategy. Suppose we are
learning the selectional preference of \textit{drink}, and the network
in Figure~4 is the knowledge base. The verb occurred with
\textit{java} and \textit{water}. This situation can be represented as
a Bayesian network. The variables \emph{ISLAND} and 
\emph{BEVERAGE} represent concepts in a semantic hierarchy.
The variables \emph{java} and \emph{water} stand for possible
instantiations of the concepts.
All the variables are Boolean; i.e., they are associated with two
states, \textit{true} or \textit{false}. Suppose the following CPTs
define the priors associated with each node.\footnote{\textit{I, B, j} and \textit{w}
abbreviate \textit{ISLAND, BEVERAGE, java and water}, respectively.} The
unconditional probabilities are 
$P(I=true)=P(B=true)=0.01$ and $P(I=false)=P(B=false)=0.99$,
and the CPTs for the child nodes are
\begin{center}
\begin{footnotesize}
\begin{tabular}{|l|c|c|c|c|} \hline
 & \multicolumn{4}{c|}{$P(X=x|Y_1 = y_1,Y_2=y_2)$} \\ \hline \hline
 & $I,B$ & $I,\neg B$ & $\neg I,B$ & $\neg I,\neg B$ \\ \hline 
 \textit{j = true} & 0.99 & 0.99 & 0.99 & 0.01 \\
 \textit{j = false} & 0.01 & 0.01 & 0.01 & 0.99 \\ \hline \hline
 \textit{w = true} & 0.99 & 0.99 & 0.01 & 0.01  \\
 \textit{w = false} & 0.01 & 0.01 & 0.99 & 0.99 \\ \hline
\end{tabular}
\end{footnotesize}
\end{center}
These values mean that the occurrence of either concept is
\emph{a priori} unlikely. If either concept is true the word
\emph{java} is likely to occur. Similarly, if \emph{BEVERAGE} occurs it
is likely to observe also the word \emph{water}. As the posterior
probabilities show, if \emph{java}
occurs, the beliefs in both concepts increase: \(P(I|j) = P(B|j) =
0.3355\). However, \emph{water} 
provides evidence for \emph{BEVERAGE} only. Overall there is more evidence
for the hypothesis that the concept being expressed is
\emph{BEVERAGE} and not \emph{ISLAND}. Bayesian networks implement
this inference scheme; if we compute the 
conditional probabilities given that both words occurred, we obtain
\(P(B|j,w) = 0.98\) and \(P(I|j,w) = 0.02\).
The new evidence caused the ``island'' hypothesis to be \textit{explained
away}!

\subsection{The relevance of priors}
Explaining away seems to depend on the specification of the prior
probabilities. The priors define the background knowledge available to
the model relative to the conditional probabilities of the events
represented by the variables, but also about the joint distributions
of several events. In the simple network above, we defined the
probability that either concept is selected (i.e., that the
corresponding variable is true) to be extremely 
small. Intuitively, there are many concepts and the probability
of observing any particular one is small. 
This means that the joint probability of the two events is much
higher in the case in which only one of them is true (0.0099) than in
the case in which they are both true (0.0001). Therefore, via the priors,
we introduced a bias according to which the hypothesis that one
concept is selected will be
favored over two co-occurring ones. This is a general pattern of
Bayesian networks; the prior causes simpler explanations to be
preferred over more complex ones, and thereby the explaining away effect.

\section{A Bayesian network approach to learning selectional preference}

\subsection{Structure and parameters of the model}
\begin{figure}
\begin{center}
\epsfig{file=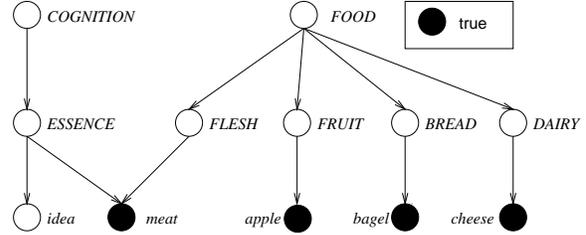, width = 3in}
\caption{A Bayesian network for the simple example.}
\end{center}
\end{figure}

The hierarchy of nouns in Wordnet defines a DAG. Its mapping
into a BBN is straightforward. Each word or synset in Wordnet is a
\textbf{node} in the network. If A is a hyponym of B there is an
\textbf{arc} in the network from B to A. All the variables are
Boolean. A synset node is \textit{true} if the verb selects
for that class. A word node is \textit{true} if the word can appear as
an argument of the verb.
The priors are defined following two intuitive principles. First,
it is \textit{unlikely} that a verb \emph{a priori} selects for
any particular synset. Second, if a verb does
select for a synset, say \textit{FOOD}, then it is \textit{likely} 
that it also selects for its hyponyms, say \textit{FRUIT}.
The same principles apply to words: it is \textit{likely} that a word
appears as an argument of the verb if the verb selects for any of its
possible senses. On the other hand, if the verb does not select
for a synset, it is \textit{unlikely} that the words
instantiating the synset occur as its arguments. ``Likely'' and
``unlikely'' are given numerical values that sum up to 1.
The following table defines the scheme for the CPTs associated with
each node in the network; $p_i(X)$ denotes the $ith$ parent of the
node $X$.
\begin{center}
\begin{footnotesize}
\begin{tabular}{|l|c|} \hline
 & $P(X=x|p_1(X)\vee,\ldots,\vee p_n(X) = true)$ \\ \hline \hline
 \textit{x = true} & \textit{likely} \\ \hline
 \textit{x = false} & \textit{unlikely} \\ \hline \hline
 & $ P(X=x|p_1(X)\wedge,\ldots,\wedge p_n(X) = false) $ \\ \hline \hline
\textit{x = true} & \textit{unlikely} \\ \hline
\textit{x = false}& \textit{likely} \\ \hline
\end{tabular}
\end{footnotesize}
\end{center}
For the root nodes, the table reduces to the unconditional probability
of the node. Now we can test the model on the simple example seen
earlier. $W^+$ is the set of words that occurred with the verb. The nodes
corresponding to the words in $W^+$ are set to \textit{true} and the
others left unset. For the previous example $W^+ = \{meat, apple, bagel, cheese\}$,
and the corresponding nodes are set to \textit{true}, as depicted in
Figure~5. With \textit{likely} and \textit{unlikely} respectively equal to
0.99 and 0.01, the posterior probabilities
are\footnote{\textit{F, C, m, a, b} and \textit{c} respectively stand for
\textit{FOOD, COGNITION, meat, apple, bagel} and \textit{cheese}}
\(P(F|m,a,b,c) = 0.9899\) and \(P(C|m,a,b,c) = 0.0101\).
Explaining away works. The posterior probability of \textit{COGNITION}
gets as low as its prior, whereas the probability of \textit{FOOD} goes up
to almost 1. A Bayesian network approach seems to actually implement
the \textit{conservative} strategy we thought to be the correct
one for unsupervised learning of selectional restrictions.

\subsection{Computational issues in building BBNs based on Wordnet}

The implementation of a BBN for the whole of Wordnet faces
computational complexity problems typical of graphical models. 
A densely connected BBN presents two kinds of problems. The
first is the storage of the CPTs. The size of a CPT grows
exponentially with the number of parents of the node.\footnote{Some
words in Wordnet have more than 20 senses. For example, \textit{line} in
Wordnet is associated with 25 senses. The size of its CPT is
therefore $2^{26}$. Storing a table of float numbers for this node alone
requires around $(2^{26})8=537$ MBytes of memory.}
This problem can be solved by optimizing the representation of these tables.
In our case most of the entries have the same values, and a compact
representation for them can be found (much like the one used in the
\textbf{noisy-OR} model~\cite{Pearl88}). 

A harder problem is performing inference.
The graphical structure of a BBN represents the dependency relations
among the random variables of the network.
The algorithms used with BBNs usually perform inference by dynamic
programming on the triangulated moral graph.
A lower bound on the number of computations that are necessary to model the
joint distribution over the variables using such algorithms is
\(2^{|n|+1}\), where $n$ is the size of the maximal boundary set
according to the visitation schedule.

\subsection{Subnetworks and balancing}
\begin{figure}
\begin{center}
\epsfig{file=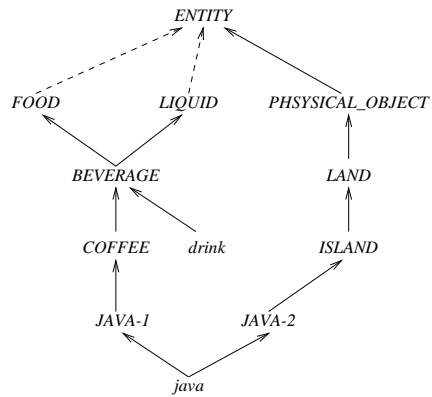, width = 2.2in}
\caption{The subnetwork for \textit{drink}.}
\end{center}
\end{figure}
Because of these problems we could not build a single BBN for Wordnet.
Instead we simplified the structure of the model by building a smaller
subnetwork for each predicate-argument pair. 
A subnetwork consists of the union of the sets of ancestors of
the words in $W^+$. Figure~6 provides an example of the union of these 
``ancestral subgraphs'' of Wordnet for the words \textit{java} and
\textit{drink} (compare it with Figure~1). 

This simplification 
does not affect the
computation of the distributions we are interested in; that is, the 
\textit{marginals} of the synset nodes. 
A BBN provides a compact representation for the joint distribution over 
the set of variables in the network. If \(N = X_1,...,X_n\) is a Bayesian
network with variables \(X_1,...,X_n\), its joint distribution
\(P(N)\) is the product of all the conditional probabilities specified 
in the network,
\begin{equation} \label{bn1}
 P(N) = \prod_{j}~P(X_j|~pa(X_j))
\end{equation}
where \(pa(X)\) is the set of parents of \(X\). A BBN generates a
factorization of the joint distribution over its variables.
Consider a network of three nodes $A, B, C$ with arcs from $A$ to $B$
and $C$. Its joint distribution can be characterized as
$P(A,B,C) = P(A)P(B|A)P(C|A)$.
If there is no evidence for $C$ the joint distribution is
\begin{eqnarray} \label{bn3}
 P(A,B,C) & = & P(A)P(B|A)\sum_{C}P(C|A)\nonumber\\
          & = & P(A)P(B|A)\nonumber \\
          & = & P(A,B)
\end{eqnarray}
The node $C$ gets marginalized out. Marginalizing over a childless node
is equivalent to removing it with its connections from the
network. Therefore the subnetworks are equivalent to the whole
network; i.e., they have the same joint distribution. 

Our model computes the value of $P(c|p,r)$, but we did not compute the
prior $P(c)$ for all nouns in the corpus. We assumed this to be a
constant, equal to the \textit{unlikely} value, for all classes.
In a BBN the values of the marginals increase with their distance from
the root nodes. To avoid undesired bias (see table of results) we
defined a balancing formula that adjusted 
the conditional probabilities of the CPTs in such a way that we 
got all the marginals to have approximately the same
value.\footnote{More details can be found in an extended version of
the paper: www.cog.brown.edu/$\sim$massi/.}

\section{Experiments and results\footnote{For these experiments we used values for the
\textit{likely} and \textit{unlikely} parameters of 0.9 and 0.1, respectively.}}
\subsection{Learning of selectional preferences}

\begin{table}
\begin{center}
\begin{small}
\begin{tabular}{|l|l|l|} \hline
 Ranking & Synset & $P(c|p,r)$ \\ \hline \hline
 1 & \textit{VEHICLE} & $0.9995$ \\ \hline
 2 & \textit{VESSEL} & $0.9893$ \\ \hline
 3 & \textit{AIRCRAFT} & $0.9937$ \\ \hline
 4 & \textit{AIRPLANE} & $0.9500$ \\ \hline
 5 & \textit{SHIP} & $0.9114$ \\ \hline
 \ldots & \ldots & \ldots \\ \hline
255 & \textit{CONCEPT} & $0.1002$ \\ \hline
256 & \textit{LAW} & $0.1001$ \\ \hline
257 & \textit{PHILOSOPHY} & $0.1000$ \\ \hline
258 & \textit{JURISPRUDENCE} & $0.1000$ \\ \hline
\end{tabular}
\caption{Results for (\textit{maneuver,object}).}
\end{small}
\end{center}
\end{table}

When trained on predicate-argument pairs extracted from a large
corpus, the San Jose Mercury Corpus, the model gave very good results. The
corpus contains about 1.3 million verb-object tokens.
The obtained rankings of classes according to their posterior
marginal probabilities were good. Table~1 shows the top and the bottom
of the list of synsets for the verb \textit{maneuver}. The model
learned that \textit{maneuver} ``selects'' for members of the class
\textit{VEHICLE} and of other plausible classes, hyponyms of
\textit{VEHICLE}. It also learned that the verb does not select for
direct objects that are members of classes, like \textit{CONCEPT} or
\textit{PHILOSOPHY}.

\subsection{Word sense disambiguation test}

A direct evaluation measure for unsupervised learning of SP models
does not exist. These models are instead evaluated on a word-sense
disambiguation test (WSD). The idea is that systems that learn SP
produce word sense disambiguation as a side-effect. 
\textit{Java} might be interpreted as the \textit{island} or the
\textit{beverage}, but in a context like ``the tourists flew to 
Java'' the former seems more correct, because \textit{fly} could
select for geographic locations but not for beverages. A system
trained on a predicate \textit{p} should be able to disambiguate 
arguments of \emph{p} if it has learned its selectional restrictions.

We tested our model using the test and training data developed by
Resnik (see Resnik, 1997). The same test was used
in~\cite{Abney99}. The training data consists of 
predicate-object counts extracted from 4/5 of the Brown corpus (about
1M words). The test set consists of predicate-object
pairs from the remaining 1/5 of the corpus, which has been
manually sense-annotated by Wordnet researchers.
The results are shown in Table~2. 
The baseline algorithm chooses at random one of the multiple senses of 
an ambiguous word. The ``first sense'' method always chooses the most
frequent sense (such a system should be trained on sense-tagged data).
\begin{table}
\begin{center}
\begin{tabular}{|l|r|} \hline
 Method & Result \\ \hline \hline
 Baseline & $28.5\% $ \\ \hline
 Abney and Light (HMM smoothed) & $35.6\% $ \\ \hline
 Abney and Light (HMM balanced) & $42.3\% $ \\ \hline
 Resnik & $44.3\% $ \\ \hline
 BBN (without balancing) & $45.6\%$ \\ \hline
 BBN (with balancing)& $51.4\% $ \\ \hline
 First Sense & $82.5\% $ \\ \hline
\end{tabular}
\caption{Results}
\end{center}
\end{table}
Our model performed better than the state of the art models for
unsupervised learning of SP. It seems to define a better estimator for
\(P(c|p,r)\).

It is remarkable that the model achieved
this result making only a limited use of distributional information. 
A noun is in $W^+$ if it occurred at least once in the
training set, but the system does not know if it occurred once or
several times; either it occurred or it didn't. The model did not
suffer too much from this limitation during this task. This is
probably due to the sparseness of the training data for the test.
For each verb the average number of object types is $3.3$, for each of 
them the average number of tokens is $1.3$; i.e., most of the
words in the training data only occurred once. For this training set
we also tested a version of the model that built a word node for each
observed object token and therefore integrated the distributional
information. On the WSD test it performed exactly the same as the
simpler version.
When trained on the San Jose Mercury Corpus the model performed worse
on the WSD test ($35.8\%$). This is not too surprising
considering the differences between the SJM and the Brown corpora: the 
former, a recent
newswire corpus; the latter, an older, balanced corpus. Another
important factor is the different relevance 
of distributional information. The training data from the SJM Corpus
are much richer and noisier than the Brown data. Here the
frequency information is probably crucial; however, in this case we
could not implement the simple scheme above. 

\subsection{Conclusion}
Explaining away implements a cognitively attractive and successful
strategy. A straightforward improvement would be for the model to make
full use of the distributional information present in the training data;
we only partially achieved this.
Bayesian networks are usually confronted with a single presentation of
evidence. Their extension to multiple evidence is not trivial. We
believe the model can be extended in this direction. Possibly
there are several ways to do so (multinomial sampling,
dedicated implementations, etc.). However, we believe that the most
relevant finding of this research might be that ``explaining away'' is
not only a property of Bayesian networks but of Bayesian inference in
general and that it might be implementable in other kinds of
graphical models. We observed that the property seems to depend on the
specification of the \textit{prior probabilities}. We found that 
the HMM model of~\cite{Abney99} was \textit{unidentifiable}; that is,
there are several solutions for the parameters of the model, including 
the desired one. Our intuition is that it should be possible to
implement ``explaining away'' in a HMM with priors, so that it would
prefer only one or a few solutions over many. This model would have also
the advantage of being computationally simpler. 

\bibliography{selpref} 
\end{document}